%% file: main.tex
\documentclass{article}

\usepackage{spconf}
\usepackage{times}
\usepackage{epsfig}
\usepackage{graphicx}
\usepackage{amsmath}
\usepackage{amssymb}

\usepackage{siunitx}
\usepackage{booktabs}
\usepackage{paralist}

\usepackage[pagebackref=true,breaklinks=true,letterpaper=true,colorlinks,bookmarks=false]{hyperref}

\DeclareGraphicsExtensions{.pdf,.jpg,.png}
\graphicspath{{figs/}}

\newcommand{\secref}[1]{Section~\ref{sec:#1}}
\newcommand{\figref}[1]{Figure~\ref{fig:#1}}
\newcommand{\tblref}[1]{Table~\ref{tbl:#1}}

\title{Single Image Cloud Detection via Multi-Image Fusion}

\name{Scott Workman$^1$ \hspace{.4cm} M. Usman Rafique$^2$  \hspace{.4cm} Hunter Blanton$^2$ \hspace{.4cm} Connor Greenwell$^2$  \hspace{.4cm} Nathan Jacobs$^2$}
\address{$^1$DZYNE Technologies \hspace{1.9in} $^2$University of Kentucky}

\begin{document}

\maketitle

\begin{abstract}
    Artifacts in imagery captured by remote sensing, such as clouds, snow, and shadows, present challenges for various tasks, including semantic segmentation and object detection. A primary challenge in developing algorithms for identifying such artifacts is the cost of collecting annotated training data. In this work, we explore how recent advances in multi-image fusion can be leveraged to bootstrap single image cloud detection. We demonstrate that a network optimized to estimate image quality also implicitly learns to detect clouds. To support the training and evaluation of our approach, we collect a large dataset of Sentinel-2 images along with a per-pixel semantic labelling for land cover. Through various experiments, we demonstrate that our method reduces the need for annotated training data and improves cloud detection performance.
\end{abstract}

\input{1_intro.tex}
\input{2_approach.tex}
\input{3_experiments.tex}
\input{4_conclusion.tex}

{\small
\bibliographystyle{IEEEbib}
\bibliography{biblio}
}

\end{document}

%% file: 1_intro.tex
\section{Introduction}

As overhead imagery captured via remote sensing becomes more abundant,
it is increasingly relied upon as an important source of information
for understanding locations and how they change over time. For
example, methods have been proposed for extracting
roads~\cite{batra2019improved}, detecting
buildings~\cite{bischke2019multi}, estimating land
cover~\cite{robinson2019large}, and interpreting the effects of
natural disasters~\cite{doshi2018satellite}. Unfortunately, various
artifacts contained in the captured imagery, such as clouds, snow, and
shadows, negatively impact the performance of these methods. 

Clouds and their properties have long been researched due to their
impact on weather and climate processes~\cite{liou1986influence}. In
an empirical study Wylie et al.~\cite{wylie2005trends} analyze cloud
cover over a 22 year period using atmospheric sounding, finding that
approximately 75 percent off all observations indicated clouds. Given
their high frequency, clouds present persistent challenges for
interpreting overhead imagery and many methods have been proposed for
identifying them~\cite{li2015cloud,xie2017multilevel}. 

The primary challenge is that the appearance of clouds can vary
dramatically and collecting manually annotated data is time consuming
and expensive. This issue is further compounded by the various sensor
types and resolutions of satellite imagery, as well as differences in
locations around the globe. Consider the scenario of transitioning to
a new sensor. Instead of collecting large amounts of new annotations,
a method is needed that can function with minimal supervision. In this
work we explore how recent advances in multi-image fusion can be
extended to support cloud detection. 

First, we design an architecture for weakly-supervised multi-image
fusion that learns to estimate image quality. Then, we describe two
approaches which take advantage of the resulting quality network to
produce a cloud detector. To support the training and evaluation of
our methods, we collect a large dataset of overhead images captured at
varying timesteps and varying levels of cloud cover. Our contributions
include: 1) an analysis of multi-image fusion on real data, 2) two
approaches for identifying clouds that require limited supervision,
and 3) an extensive evaluation, achieving state-of-the-art results on
a benchmark dataset.

%% file: 2_approach.tex
\section{Approach}

Our approach for identifying clouds uses multi-image fusion as a form
of bootstrapping, reducing the need for annotated training data. We
start by describing the architecture for multi-image fusion and then
describe how we extend this architecture for detecting clouds.

\subsection{Multi-Image Fusion}
\label{sec:fusion}

We apply multi-image fusion to take a stack of images over the same
region, $I = \{I_1,\ldots,I_K\}$, where $I_j \in R^{h \times w \times
3}$, and produce a fused image, $F = \phi(I)$, such that $F$ is free
of artifacts. Our approach is inspired by the recent work of Rafique
et al.~\cite{usman2019weakly}. There are two main steps: 1) estimating
a per-pixel quality mask for each image then using the qualities to
compute a fused image and 2) passing the fused image through a
segmentation network to produce a per-pixel semantic labeling. When
trained end-to-end, this architecture learns to estimate per-pixel
image qualities that can be used to produce a fused image with reduced
artifacts, without requiring explicit labels.

\subsubsection{Dataset}

\begin{figure}[t]
  \centering
  \setlength\tabcolsep{1pt}
    \begin{tabular}{ccc}
        \includegraphics[width=.32\linewidth]{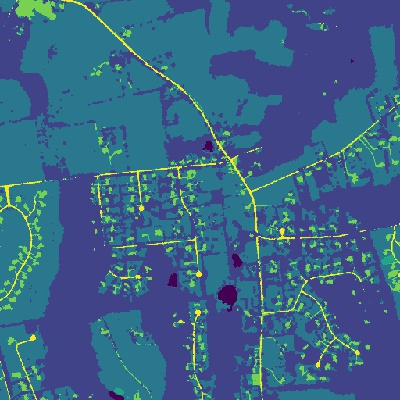} &
        \includegraphics[width=.32\linewidth]{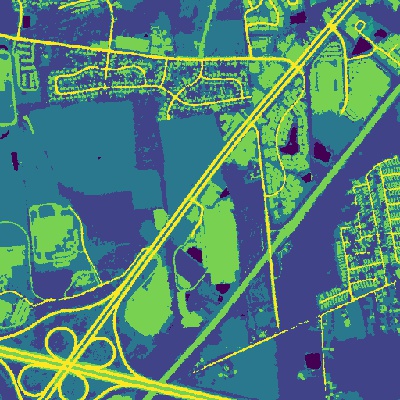} &
        \includegraphics[width=.32\linewidth]{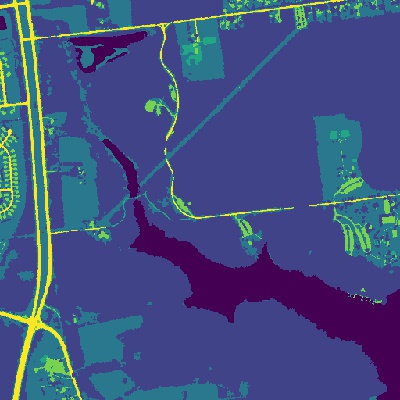} \\
        
        \includegraphics[width=.32\linewidth]{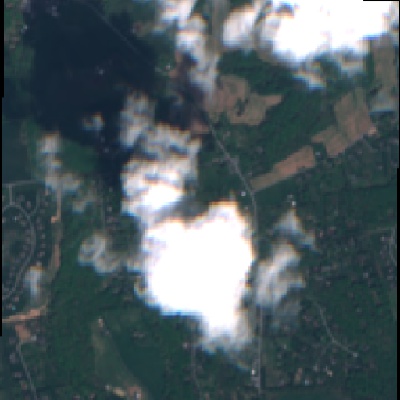} &
        \includegraphics[width=.32\linewidth]{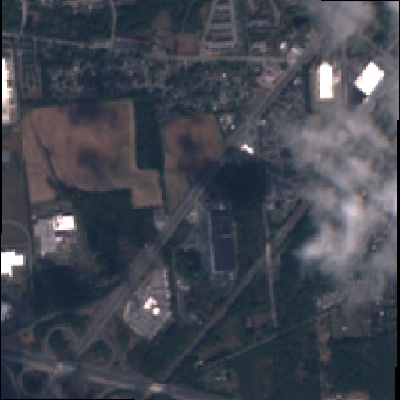} &
        \includegraphics[width=.32\linewidth]{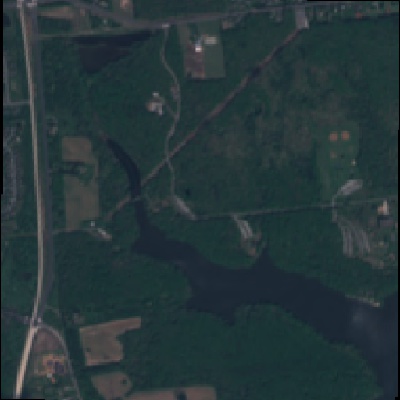} \\
        
        \includegraphics[width=.32\linewidth]{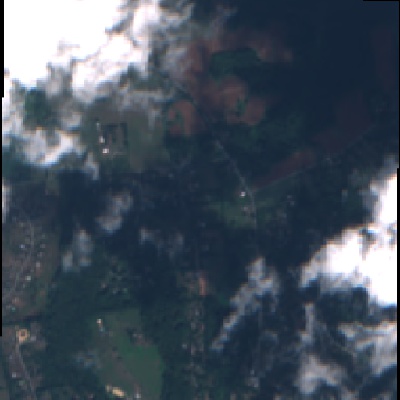} &
        \includegraphics[width=.32\linewidth]{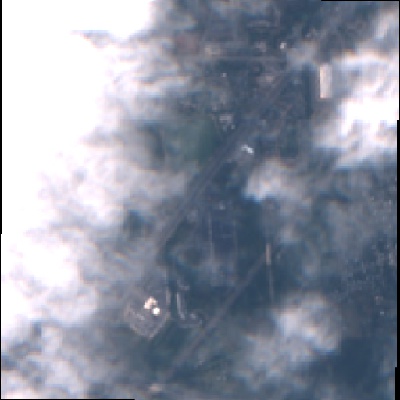} &
        \includegraphics[width=.32\linewidth]{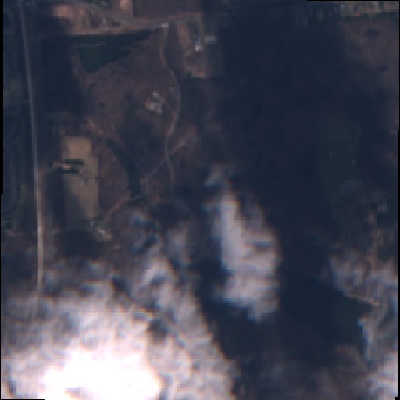} \\
        
        \includegraphics[width=.32\linewidth]{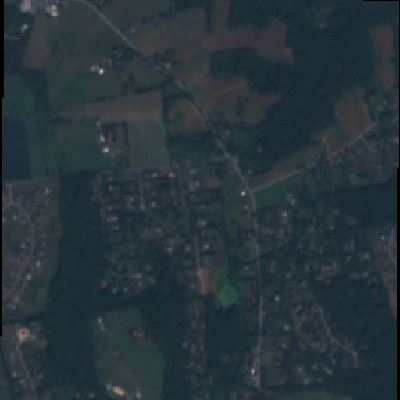} &
        \includegraphics[width=.32\linewidth]{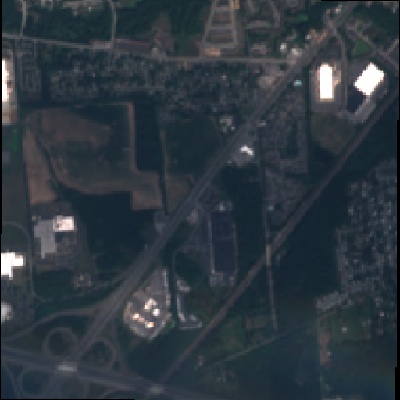} &
        \includegraphics[width=.32\linewidth]{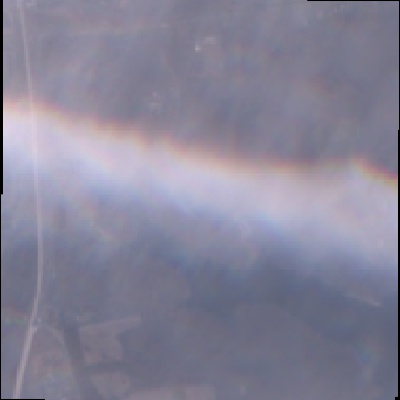} \\
    \end{tabular}

    \caption{Examples from our dataset for multi-image fusion. (top)
    Land cover labeling from the Chesapeake Land Cover
    dataset~\cite{robinson2019large}. (bottom) Images of the same
    location with varying cloud cover.}

  \label{fig:data}
\end{figure}

To support the training of our methods, we collected Sentinel-2
imagery from the state of Delaware with varying levels of cloud cover.
Starting from a bounding box around the state, we generated a set of
non-overlapping tiles using the standard XYZ style spherical Mercator
tile. For each tile, we collected a semantic labeling from the
Chesapeake Land Cover dataset~\cite{robinson2019large}, removing tiles
without valid labels. For each remaining tile, we randomly downloaded
six Sentinel-2 images (RGB bands) from the year 2019 that satisfied
the constraint of having between 10\% and 50\% cloud cover in the
parent Sentinel-2 image strip. This process resulted in \num{1033}
unique locations and \num{6198} images (of size $512 \times 512$).
\figref{data} shows some example images from our dataset.

\subsubsection{Method}

Each image $I_j$ is first passed through a \emph{quality} network
which outputs a per-pixel quality mask $Q_j \in R^{h \times w}$ for
each pixel $p$, such that $Q_j(I_j(p)) \in [0,1]$. Given quality masks
for each image, a relative quality score at each pixel is computed by
applying a softmax across images:
\begin{equation}
    Q^{*}_j(p) = \frac{e^{Q_j(p)}}{\sum_{k=1}^{K} e^{Q_k(p)}}.
\end{equation}
The final fused image $F_j$ is obtained by averaging the images
weighted by the relative quality score:
\begin{equation}
    F_{j}(p) = \sum_{j=1}^{K} I_j(p) Q^{*}_j(p).
\end{equation}
The fused image $F_j$ is passed through a \emph{segmentation} network
to produce a per-pixel labeling. The entire architecture, both quality
network and segmentation network, are optimized using a cross-entropy
loss function.

\subsubsection{Architecture Details}

For the quality network, we use a slightly modified
U-Net~\cite{ronneberger2015u} with the same number of layers but a
quarter of the feature maps compared to the original work. The final
activation is a sigmoid. For the segmentation network, we build on
LinkNet~\cite{chaurasia2017linknet}, a modern, lightweight
segmentation architecture that follows an encoder/decoder approach.
Specifically, we use LinkNet-34, which is LinkNet with a
ResNet-34~\cite{he2016deep} encoder. We initialize the encoder with
weights from a network pretrained on ImageNet. 

\subsection{Detecting Clouds}

\begin{figure*}
  \centering
  \setlength\tabcolsep{1pt}
    \begin{tabular}{cccc|ccc}
        Image (1 of 6) & Quality & Image (2 of 6) & Quality & Fused & Target & Prediction \\
        
        \includegraphics[width=.139\linewidth]{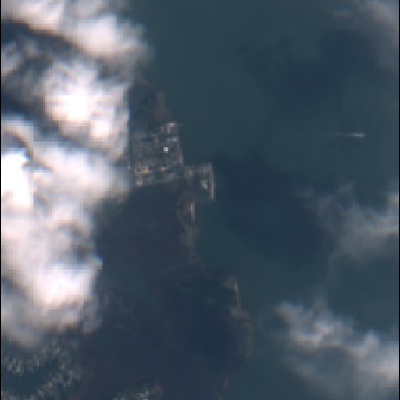} &
        \includegraphics[width=.139\linewidth]{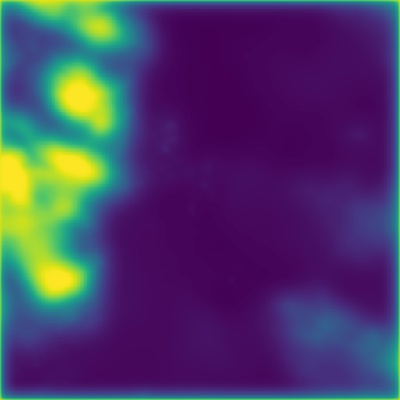} &
        \includegraphics[width=.139\linewidth]{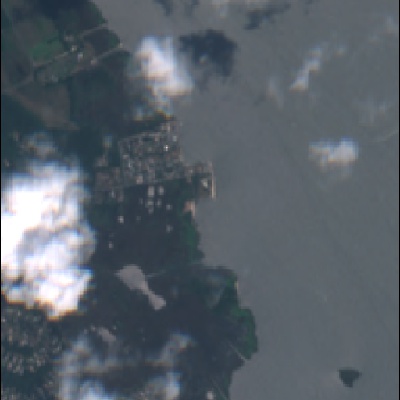} &
        \includegraphics[width=.139\linewidth]{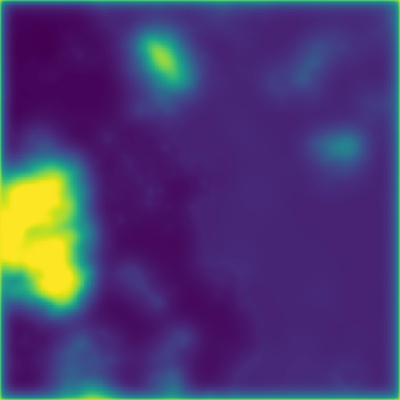} &
        \includegraphics[width=.139\linewidth]{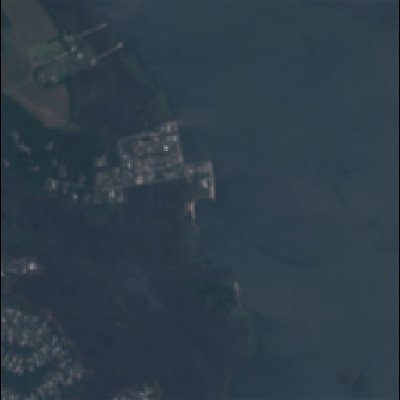} &
        \includegraphics[width=.139\linewidth]{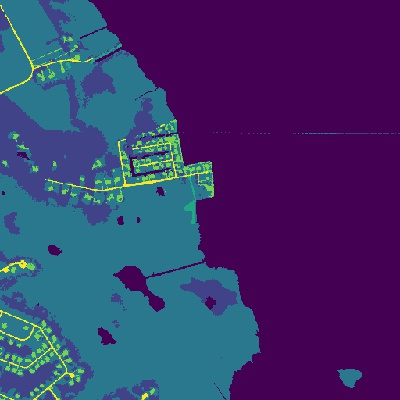} &
        \includegraphics[width=.139\linewidth]{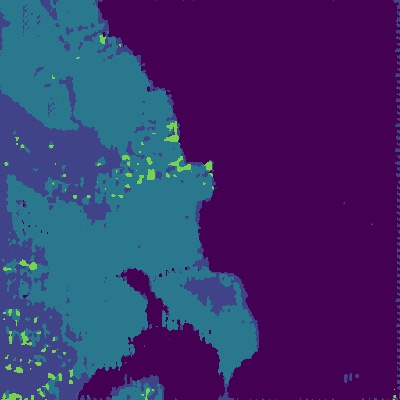} \\
        
        \includegraphics[width=.139\linewidth]{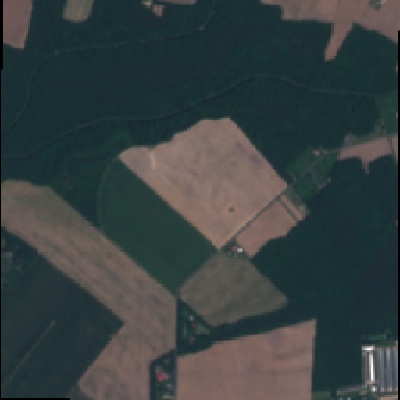} &
        \includegraphics[width=.139\linewidth]{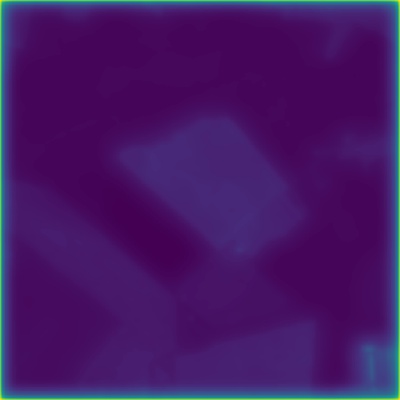} &
        \includegraphics[width=.139\linewidth]{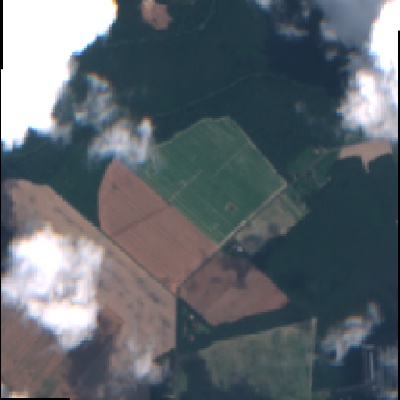} &
        \includegraphics[width=.139\linewidth]{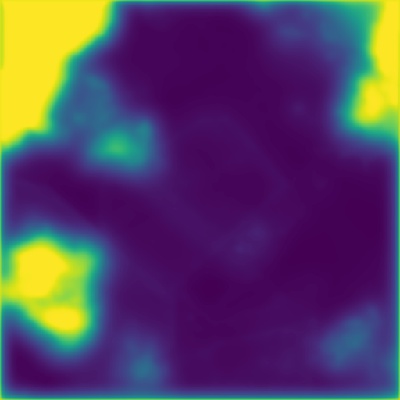} &
        \includegraphics[width=.139\linewidth]{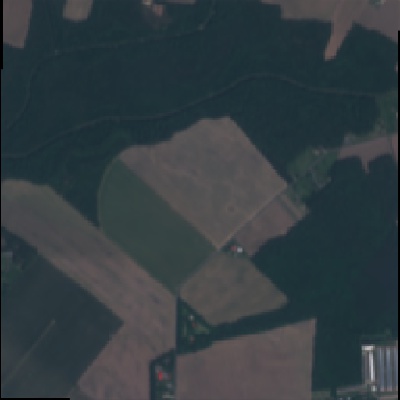} &
        \includegraphics[width=.139\linewidth]{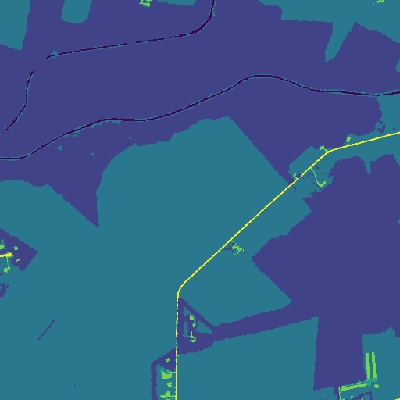} &
        \includegraphics[width=.139\linewidth]{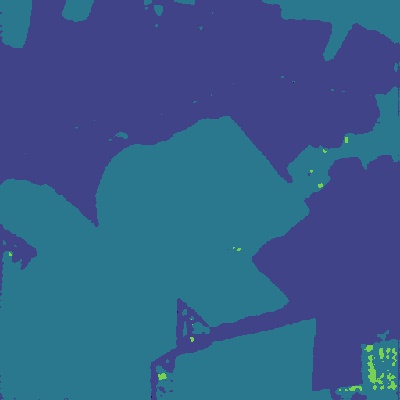} \\

    \end{tabular}

    \caption{Qualitative examples of multi-image fusion. (left)
    Example images and estimated quality masks. (right) The fused
    image produced using the relative quality scores, the target
    segmentation mask, and our prediction.}

  \label{fig:fusion}
\end{figure*}

The quality network learns to identify artifacts in the training data
that negatively impact the final segmentation, for example clouds and
regions of no data. We describe two approaches which use the quality
network, trained for multi-image fusion, as a starting point for
learning a cloud detector (per-pixel binary classification). For these
methods, we use the dataset recently introduced by Liu et
al.~\cite{liu2019clouds} with 100 training images and 20 testing
images.

\subsubsection{Quality Calibration}
We apply Platt scaling (which we refer to as quality calibration) to
transform the outputs of the quality network into a distribution over
classes (cloud/not cloud). In practice, this means we fit a logistic
regression model:
\begin{equation}
    P(y=1|Q_j(p)) = \frac{1}{1 + e^{\beta_0Q_j(p)+\beta_1}},
\end{equation}
where $\beta_0$ and $\beta_1$ are two learned parameters.

\subsubsection{Fine-Tuning the Quality Network}
Alternatively, we employ transfer learning, freezing all layers of the
quality network except the final three convolutional layers (the last
upsampling block and the final $1\times1$ convolution). Then, we
fine-tune the network for cloud detection. We optimize the network
using the following loss function:
\begin{equation}
    \mathcal{L} = \mathcal{L}_{bce} + (1 -\mathcal{L}_{dice})
\end{equation}
where $\mathcal{L}_{bce}$ is binary cross entropy, a standard loss
function used in binary classification tasks, and
$\mathcal{L}_{dice}$ is the dice coefficient, which measures spatial
overlap. 

\subsection{Implementation Details}

Our methods are implemented using the
PyTorch~\cite{paszke2017automatic} framework and optimized using
RAdam~\cite{liu2019variance} with Lookahead~\cite{zhang2019lookahead}
($k=5, \alpha=.5$). The learning rate is $\lambda = 10^{-4}$
($10^{-2}$ when fine-tuning). We train all networks with a batch size
of 10 for 100 epochs and train on random crops of size $416 \times
416$. For multi-image fusion, we randomly sample 4 images per location
during training.

%% file: 3_experiments.tex
\section{Evaluation}

We evaluate our methods both qualitatively and quantitatively through
a variety of experiments.

\subsection{Visual Analysis of Multi-Image Fusion on Real Data}

Previous work on multi-image fusion used training data augmented with
synthetic clouds~\cite{usman2019weakly}. In our work, we train and
evaluate our approach using real images with varying levels of cloud
cover. \figref{fusion} shows example output from our network
(described in \secref{fusion}), including: example images alongside
the estimated quality mask, the fused image using the relative quality
scores, the target label from the Chesapeake Land Cover
dataset~\cite{robinson2019large}, and our prediction. The estimated
quality masks clearly identify artifacts in the imagery, such as
clouds.

\subsection{Quantitative Analysis of Cloud Detection}

\begin{figure}[t]
    \centering
    \includegraphics[width=1\linewidth]{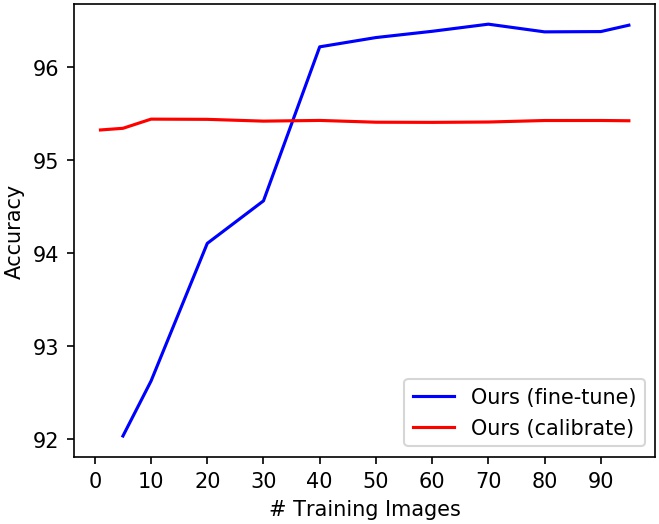}
    \caption{Evaluating the impact of the number of training images on
    cloud detection accuracy.}
    \label{fig:acc_vs_size}
\end{figure}

\begin{figure}
    \centering
    \setlength\tabcolsep{1pt}
    \begin{tabular}{cccc}
        Image & Target & Prediction & Error \\
        \includegraphics[width=.24\linewidth]{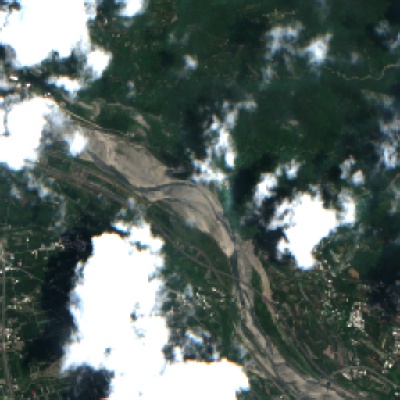} &
        \includegraphics[width=.24\linewidth]{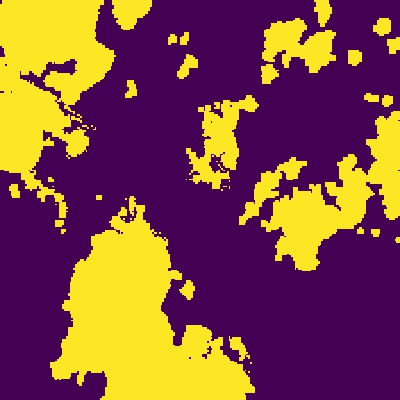} &
        \includegraphics[width=.24\linewidth]{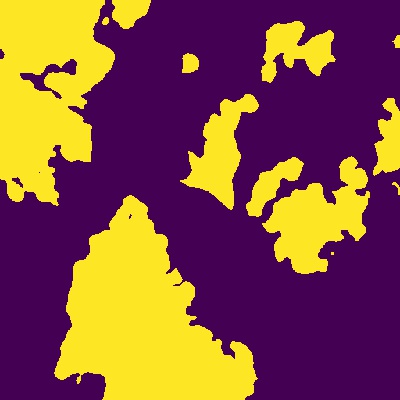} &
        \includegraphics[width=.24\linewidth]{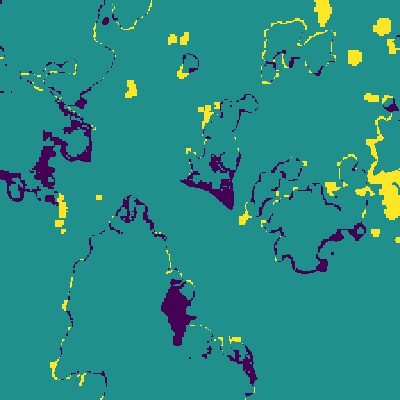} \\
        
        \includegraphics[width=.24\linewidth]{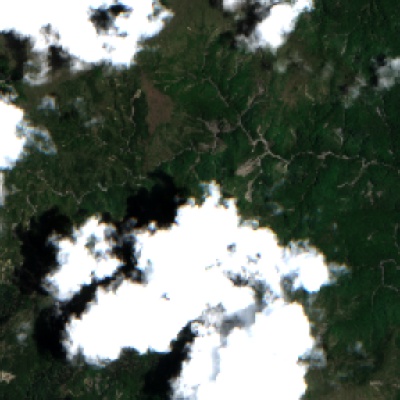} &
        \includegraphics[width=.24\linewidth]{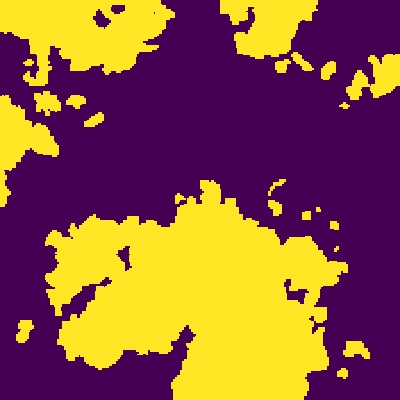} &
        \includegraphics[width=.24\linewidth]{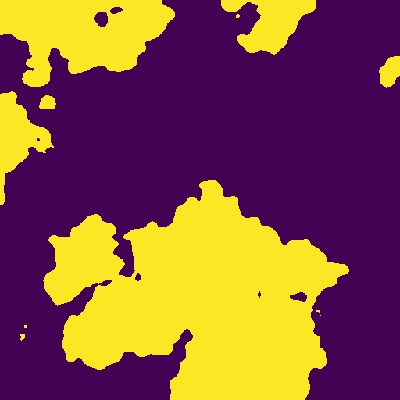} &
        \includegraphics[width=.24\linewidth]{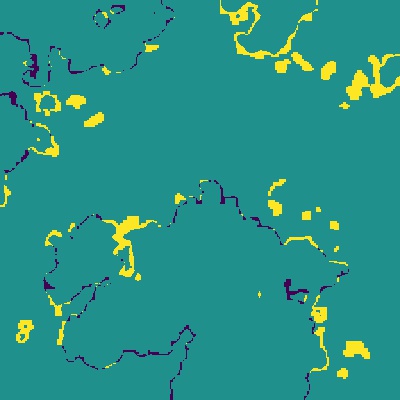} \\
        
        \includegraphics[width=.24\linewidth]{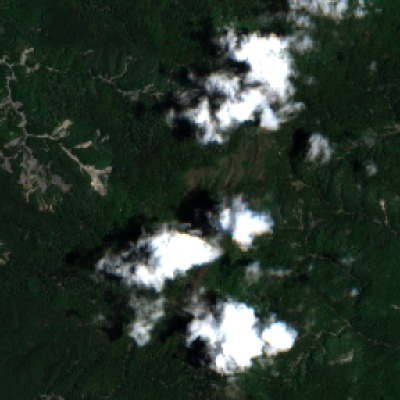} &
        \includegraphics[width=.24\linewidth]{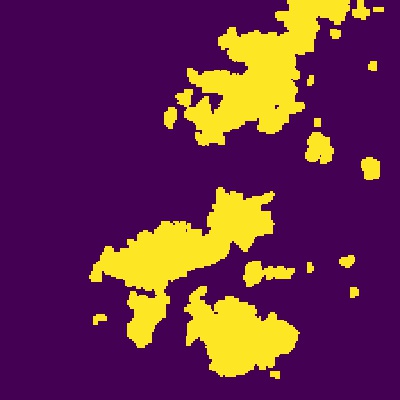} &
        \includegraphics[width=.24\linewidth]{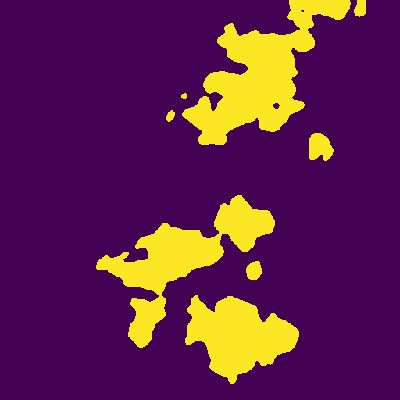} &
        \includegraphics[width=.24\linewidth]{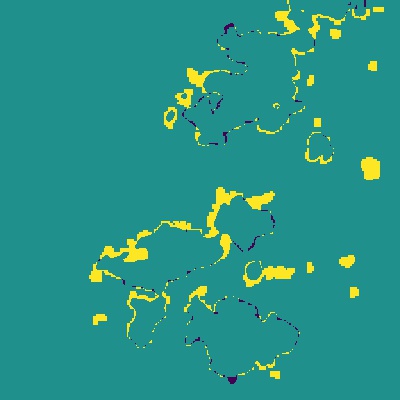} \\
    \end{tabular}
    
    \caption{Example cloud detection results using \emph{Ours
    (fine-tune)}. The error image (right) shows false positives
    (negatives) color coded as purple (yellow).}

  \label{fig:clouds}
\end{figure}

Using the dataset recently introduced by Liu et
al.~\cite{liu2019clouds}, we quantitatively evaluate our methods
ability to detect clouds. \tblref{results} shows the results of this
experiment. We compare against a baseline, \emph{Ours (threshold)},
that na\"ively thresholds the quality masks at $.5$ (treating anything
below the threshold as a cloud). The baseline, which requires no
direct supervision, is able to correctly classify over 91\% of pixels.
Applying quality calibration, \emph{Ours (calibrate)}, to the output
of the quality network improves upon this result. Ultimately
fine-tuning, \emph{Ours (fine-tune)}, outperforms all baselines,
achieving superior results than Liu et al.~\cite{liu2019clouds}. Next,
we evaluate the ability of our approach to identify clouds with
varying number of training images (\figref{acc_vs_size}). For this
experiment, we trained each model on a randomly selected subset of the
training data and fine-tuning was limited to 30 epochs. As observed,
our proposed approaches require very few annotated images to produce
reasonable cloud detection results. Finally, \figref{clouds} shows
some example predictions using our best method.

\begin{table}
  \centering
  \caption{Quantitative evaluation for cloud detection.}
  \begin{tabular}{@{}lrrrrr@{}}
    \toprule
    Method & TPR & TNR & mIoU & Accuracy \\
    \bottomrule
    Liu et al.~\cite{liu2019clouds} & 0.963 & 0.945 & 89.47\% & 95.87\% \\
    Ours (threshold) & \textbf{0.982} & 0.878 & 81.78\% & 91.73\% \\
    Ours (calibrate) & 0.933 & \textbf{0.967} & 88.50\% & 95.42\% \\
    Ours (fine-tune) & 0.962 & \textbf{0.967} & \textbf{91.24\%} & \textbf{96.51\%} \\
    \bottomrule
  \end{tabular}
  \label{tbl:results}
\end{table}

%% file: 4_conclusion.tex
\section{Conclusion}

We presented methods for detecting clouds that require minimal
supervision. Our key insight was to take advantage of multi-image
fusion, which learns to capture the notion of image quality, as a form
of pretraining. To support our methods, we introduced a large dataset
of images with varying levels of cloud cover and a corresponding
per-pixel land cover labelling. Using this dataset, we showed results
for multi-image fusion on real-world imagery. Finally, we presented a
quantitative evaluation of cloud detection, ultimately achieving
state-of-the-art results on an existing cloud detection benchmark
dataset.